%% file: paper.tex
\documentclass{article}
\usepackage{iclr}
\usepackage{amsmath}
\usepackage{amsfonts}
\usepackage{bm}
\usepackage{color}
\usepackage{colortbl}
\usepackage[hidelinks]{hyperref}
\usepackage{graphicx}
\usepackage{subcaption}
\usepackage{algorithm}
\usepackage{algorithmic}
\usepackage{adjustbox}
\usepackage{tabularx}
\usepackage{booktabs}
\usepackage{multirow}
\usepackage{cleveref}
\usepackage{tikz}

\definecolor{Gray}{gray}{0.9}

\graphicspath{{figures/}}
\usetikzlibrary{shapes.misc}

\newcommand{\iclryear}{2019}

\newcommand{\eg}{\textit{e.g\@.}}

\newcommand{\ie}{\textit{i.e\@.}}

\newcommand{\email}[1]{\href{mailto:#1}{\texttt{#1}}}

\newcommand\unitk{\,\text{k}}
\newcommand\unitM{\,\text{M}}
\newcommand\unitG{\,\text{G}}

\newcommand{\ordinal}[1]{{#1}^\mathrm{th}}
\newcommand{\realset}{\mathbb{R}}
\newcommand{\tx}{{\mathbf{x}}}

\newcommand{\tz}{{\mathbf{z}}}
\newcommand{\ti}{{\mathbf{i}}}
\newcommand{\tmu}{\bm\mu}
\newcommand{\tsigma}{\bm\sigma}
\newcommand{\convweight}{\bm\theta}
\newcommand{\bnscale}{\bm\gamma}
\newcommand{\bnshift}{\bm\beta}
\newcommand{\gateweight}{\bm\phi}
\newcommand{\gatebias}{\bm\rho}
\newcommand{\norm}{\mathsf{norm}}
\newcommand{\conv}{\mathsf{conv}}
\newcommand{\expectation}{\mathbb{E}}
\newcommand{\wta}{\mathsf{wta}}
\newcommand{\subs}{\mathsf{ss}}
\newcommand{\subsubs}{\mathsf{s}\hspace{-0.15em}}
\DeclareRobustCommand{\sparseblock}{%
  \begingroup\normalfont
  \includegraphics[height=\fontcharht\font`\B]{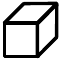}%
  \endgroup
}

\newcommand{\mcnet}{M-CifarNet}
\newcommand{\cifarx}{CIFAR-10}

\newcommand{\ilsvrc}{ILSVRC2012}
\newcommand{\resnetxviii}{ResNet-18}
\newcommand{\vggxvi}{VGG-16}

\newcommand*\samethanks[1][\value{footnote}]{\footnotemark[#1]}

\iclrfinalcopy
\title{Dynamic Channel Pruning: \\Feature Boosting and Suppression}
\author{%
    Xitong Gao$^1$\thanks{Equal contribution, corresponding authors.},
    Yiren Zhao$^2$\samethanks,
    {\L}ukasz Dudziak$^3$,
    Robert Mullins$^4$,
    Cheng-zhong Xu$^5$ \\
    $^{1, 5}$ Shenzhen Institutes of Advanced Technology, Shenzhen, China \\
    $^{2, 3, 4}$ University of Cambridge, Cambridge, UK \\
    $^1$ \email{xt.gao@siat.ac.cn},
    $^2$ \email{yaz21@cam.ac.uk}
}

\begin{document}

\maketitle

\input{abstract}
\input{intro}
\input{related}
\input{method}
\input{results}
\input{conclusion}
\input{acknowledgements}

\bibliographystyle{iclr}
\bibliography{references}

\appendix
\input{appendix}

\end{document}

%% file: abstract.tex
\begin{abstract}
Making deep convolutional neural networks
more accurate
typically comes at the cost of
increased computational and memory resources.
In this paper,
we reduce this cost
by exploiting the fact that
the importance of features
computed by convolutional layers
is highly input-dependent,
and propose feature boosting and suppression (FBS),
a new method to predictively
amplify salient convolutional channels
and skip unimportant ones at run-time.
FBS introduces small auxiliary connections
to existing convolutional layers.
In contrast to channel pruning methods
which permanently remove channels,
it preserves the full network structures
and accelerates convolution
by dynamically skipping
unimportant input and output channels.
FBS-augmented networks are trained
with conventional stochastic gradient descent,
making it readily available
for many state-of-the-art CNNs.
We compare FBS to a range of existing
channel pruning and dynamic execution schemes
and demonstrate large improvements
on ImageNet classification.
Experiments show that
FBS can respectively provide $5\times$ and $2\times$
savings in compute on \vggxvi{} and \resnetxviii{},
both with less than $0.6\%$ top-5 accuracy loss.
\end{abstract}

%% file: intro.tex
\section{Introduction}\label{sec:intro}

State-of-the-art vision and image-based tasks such as
image classification \citep{alexnet, vgg, resnet},
object detection \citep{fasterrcnn, huang17} and
segmentation \citep{long15}
are all built upon
deep \emph{convolutional neural networks} (CNNs).
While CNN architectures
have evolved to become more efficient,
the general trend has been to use larger models
with greater memory utilization,
bandwidth and compute requirements
to achieve higher accuracy.
% As all the filters in
% these models are applied blindly in a brute force manner,
% The increases in model size translate directly
% into higher compute requirements.
% The result is that typical models
% may consist of hundreds of millions of parameters
% and require billions of arithmetic operations
% to process a single image,
% with correspondingly high memory bandwidth and power demands.
The formidable amount of computational resources used by CNNs
present a great challenge in the deployment of CNNs
in both cost-sensitive cloud services
and low-powered edge computing applications.

\begin{figure}[ht]
    \centering%
    \newcommand{\maxmag}[1]{%
        \begin{minipage}{0.25\textwidth}%
            \centering{\textbf{#1}}
        \end{minipage}}
    \newcommand{\maxmags}[4]{\small%
        \maxmag{#1}\maxmag{#2}\maxmag{#3}\maxmag{#4}}
    \hspace{-0.05\textwidth}
    \begin{subfigure}{0.1\textwidth}
        \begin{flushright}\textbf{
            high \\ response \\ {~}\vspace{15pt} \\
            low \\ response \\ \vspace{15pt}}
        \end{flushright}
    \end{subfigure}~~%
    \begin{subfigure}{0.3\textwidth}
        \includegraphics[width=0.25\textwidth]{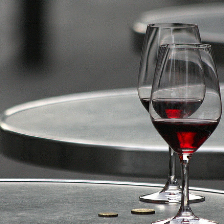}%
        \includegraphics[width=0.25\textwidth]{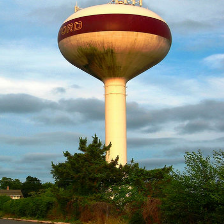}%
        \includegraphics[width=0.25\textwidth]{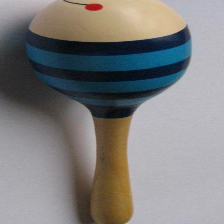}%
        \includegraphics[width=0.25\textwidth]{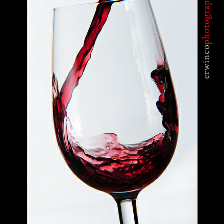}
        \maxmags{5.426}{3.297}{3.076}{2.938}\vspace{5pt}
        \includegraphics[width=0.25\textwidth]{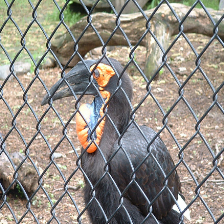}%
        \includegraphics[width=0.25\textwidth]{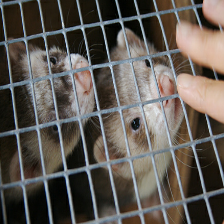}%
        \includegraphics[width=0.25\textwidth]{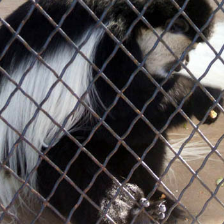}%
        \includegraphics[width=0.25\textwidth]{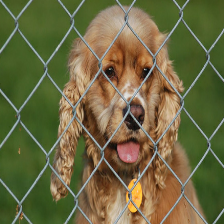}
        \maxmags{0.051}{0.052}{0.066}{0.069}
        \caption{Channel 114}\label{fig:excited:114}
    \end{subfigure}~%
    \begin{subfigure}{0.3\textwidth}
        \includegraphics[width=0.25\textwidth]{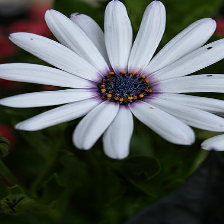}%
        \includegraphics[width=0.25\textwidth]{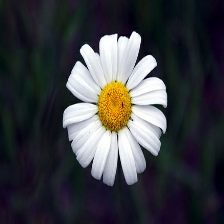}%
        \includegraphics[width=0.25\textwidth]{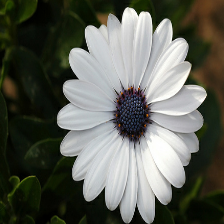}%
        \includegraphics[width=0.25\textwidth]{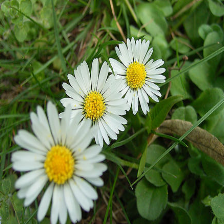}
        \maxmags{3.409}{3.309}{3.298}{3.171}\vspace{5pt}
        \includegraphics[width=0.25\textwidth]{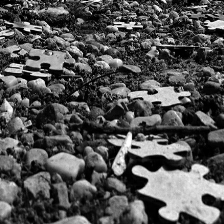}%
        \includegraphics[width=0.25\textwidth]{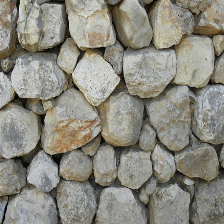}%
        \includegraphics[width=0.25\textwidth]{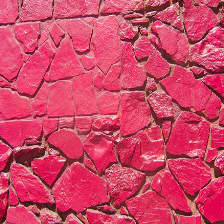}%
        \includegraphics[width=0.25\textwidth]{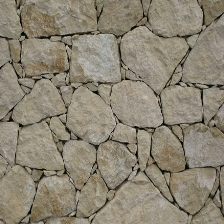}
        \maxmags{-0.229}{-0.218}{-0.168}{-0.161}
        \caption{Channel 181}\label{fig:excited:181}
    \end{subfigure}~%
    \begin{subfigure}{0.3\textwidth}
        \includegraphics[width=\textwidth]{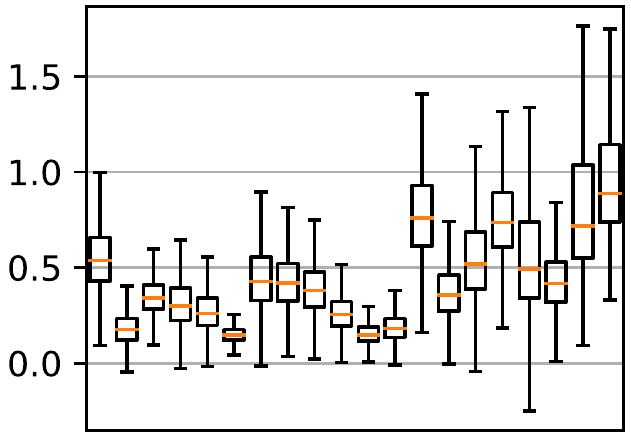}
        \caption{
            The distribution of maximum activations
        }\label{fig:excited:first20}
    \end{subfigure}
    \caption{%
        When images from the ImageNet validation dataset
        are shown to a pre-trained ResNet-18~\citep{resnet},
        the outputs from certain channel neurons
        may vary drastically.
        The top rows in
        (\subref{fig:excited:114}) and (\subref{fig:excited:181})
        are found respectively to
        greatly excite neurons in channels 114 and 181
        of layer \texttt{block\_3b/conv2},
        whereas the bottom images
        elicit little activation
        from the same channel neurons.
        The number below each image indicate
        the maximum values observed in the channel
        before adding the shortcut and activation.
        Finally, (\subref{fig:excited:first20}) shows
        the distribution of maximum activations observed
        in the first 20 channels.
    }\label{fig:excited}
\end{figure}

One common approach
to reduce the memory, bandwidth and compute costs
is to prune over-parameterized CNNs.
If performed in a coarse-grain manner this approach is
known as \emph{channel pruning}
\citep{ye18, he17, liu17, wen16}.
Channel pruning evaluates channel saliency measures
and removes all input and output connections
from unimportant channels---
generating a smaller dense model.
A saliency-based pruning method, however,
has threefold disadvantages.
Firstly, by removing channels,
the capabilities of CNNs are permanently lost,
and the resulting CNN may
never regain its accuracy for difficult inputs
for which the removed channels were responsible.
Secondly, despite the fact that
channel pruning may drastically shrink model size,
without careful design,
computational resources
cannot be effectively reduced in a CNN
without a detrimental impact on its accuracy.
Finally, the saliency of a neuron is not static,
which can be illustrated by
the feature visualization in \Cref{fig:excited}.
Here, a CNN is shown a set of input images,
certain channel neurons in a convolutional output
may get highly excited,
whereas another set of images
elicit little response from the same channels.
This is in line with our understanding of CNNs
that neurons in a convolutional layer
specialize in recognizing distinct features,
and the relative importance of a neuron
depends heavily on the inputs.

% dynamic pruning
The above shortcomings prompt the question: \emph{%
    why should we prune by static importance,
    if the importance is highly input-dependent?}
Surely, a more promising alternative is to
\emph{prune dynamically} depending on the current input.
A dynamic channel pruning strategy
allows the network to learn to
prioritize certain convolutional channels and
ignore irrelevant ones.
Instead of simply reducing model size
at the cost of accuracy with pruning,
we can accelerate convolution by
selectively computing only a subset of channels
predicted to be important at run-time,
while considering the sparse input
from the preceding convolution layer.
In effect, the amount of cached activations
and the number of read, write and arithmetic operations
used by a well-designed dynamic model
can be almost identical to
an equivalently sparse statically pruned one.
In addition to saving computational resources,
a dynamic model preserves all neurons of the full model,
which minimizes the impact on task accuracy.

In this paper, we propose
\emph{feature boosting and suppression} (FBS)
to dynamically amplify and suppress output channels
computed by the convolutional layer.
Intuitively,
we can imagine that
the flow of information of each output channel
can be amplified or restricted
under the control of a ``valve''.
This allows salient information to flow freely
while we stop all information
from unimportant channels
and skip their computation.
Unlike pruning statically,
the valves use features from the previous layer
to predict the saliency of output channels.
With conventional \emph{stochastic gradient descent} (SGD) methods,
the predictor can learn to adapt itself
by observing the input and output features
of the convolution operation.

FBS introduces tiny auxiliary connections to
existing convolutional layers.
The minimal overhead added to the existing model
is thus negligible when compared to
the potential speed up provided by the dynamic sparsity.
Existing dynamic computation strategies in CNNs
\citep{lin17, odena17, bolukbasi17}
produce on/off pruning decisions or execution path selections.
Training them thus often resorts to
reinforcement learning, which in practice
is often computationally expensive. % \citep{reinforce}
Even though FBS similarly
use non-differentiable functions,
contrary to these methods,
the unified losses are still
well-minimized with conventional SGD\@.

We apply FBS
to a custom \cifarx{} \citep{cifar} classifier
and popular CNN models such as
\vggxvi{} \citep{vgg} and \resnetxviii{} \citep{resnet}
trained on the ImageNet dataset \citep{imagenet}.
Empirical results show that
under the same speed-ups,
FBS can produce models
with validation accuracies surpassing
all other channel pruning and
dynamic conditional execution methods
examined in the paper.

%% file: related.tex
\section{Related Work}\label{sec:related}
% pruning
%   - [ ] dynamic network surgery
%   - [ ] layer-wise optimal brain damage
%   - [ ] PerforatedCNNs
%   - [x] learning the number of neurons in deep networks
%   - [x] learning to prune filters in convolutional neural networks
%   - [x] pruning filters for efficient convnets
%   - [x] soft filter pruning
%   - [x] rethinking the smaller-norm-less-informative
%   - [x] channel pruning for accelerating very deep neural networks
%   - [x] learning structured sparsity in deep neural networks
%   - [x] network slimming
% dynamic
%   - [x] blockdrop
%   - [x] sbnet
%   - [x] spatially adaptive computation time for residual networks
%   - [x] more is less: low-cost collaborative network
%   - [x] channel gating neural networks
%   - [x] changing model behavior at test-time using reinforcement learning
%   - [x] runtime neural pruning
%   - [x] dynamic capacity networks
%   - [x] adaptive neural networks for efficient inference
%   - [ ] sparsely-gated mixture-of-experts
% others
%   - multi-scale dense networks
%   - squeeze-excitation
\subsection{Structured Sparsity}

Since \cite{lecun90} introduced optimal brain damage,
the idea of creating more compact and efficient CNNs
by removing connections or neurons
has received significant attention.
Early literature on pruning deep CNNs
zero out individual weight parameters
\citep{obs94, guo2016dynamic}.
This results in
highly irregular sparse connections,
which were notoriously difficult
for GPUs to exploit.
This has prompted custom accelerator solutions
that exploit sparse weights \citep{parashar2017, han2016eie}.
Although supporting both
sparse and dense convolutions
efficiently normally involves some compromises
in terms of efficiency or performance.

Alternatively, recent work has thus increasingly
focused on introducing \emph{structured sparsity}
\citep{wen16, ye18, alvarez16, zhou16},
which can be exploited by GPUs
and allows custom accelerators
to focus solely on efficient dense operations.
\cite{wen16} added group Lasso on channel weights
to the model's training loss function.
This has the effect of reducing the magnitude of
channel weights to diminish during training,
and remove connections from zeroed-out channels.
To facilitate this process,
\cite{alvarez16} additionally used
proximal gradient descent, while
\cite{li17} and \cite{he18}
proposed to prune channels by thresholds,
\ie~they set unimportant channels to zero,
and fine-tune the resulting CNN\@.
The objective to induce sparsity in groups of weights
may present difficulties for gradient-based methods,
given the large number of weights
that need to be optimized.
A common approach to overcome this is
to solve \citep{he17} or learn \citep{liu17, ye18}
channel saliencies to drive the
sparsification of CNNs.
\cite{he17} solved an optimization problem
which limits the number of
active convolutional channels
while minimizing the reconstruction error
on the convolutional output.
\cite{liu17} used Lasso regularization
on channel saliencies to induce sparsity
and prune channels with a global threshold.
\cite{ye18} learned to sparsify CNNs
with an iterative shrinkage/thresholding algorithm
applied to the scaling factors in batch normalization.
There are methods \citep{luo17, zhuang18} that
use greedy algorithms for channel selection.
% \cite{luo17} selectively prune input channels
% while minimizing the reconstruction error
% of the convolutional output.
% \cite{zhuang18} similarly perform
% channel selection to minimize
% a joint loss of the reconstruction error
% and the classification power of convolutional channels.
\cite{huang18} and \cite{he18amc}
adopted reinforcement learning to train agents
to produce channel pruning decisions.
PerforatedCNNs, proposed by \cite{figurnov16},
use predefined masks that are model-agnostic
to skip the output pixels in convolutional layers.

\subsection{Dynamic Execution}

In a pruned model produced
by structured sparsity methods,
the capabilities of
the pruned neurons and connections
are permanently lost.
Therefore, many propose to use dynamic networks
as an alternative to structured sparsity.
During inference,
a dynamic network can use the input data
to choose parts of the network to evaluate.

Convolutional layers are usually spatially sparse,
\ie~their activation outputs may
contain only small patches of salient regions.
A number of recent publications
exploit this for acceleration.
\cite{dong17} introduced low-cost collaborative layers
which induce spatial sparsity in cheap convolutions,
so that the main expensive ones
can use the same sparsity information.
\cite{figurnov17} proposed
spatially adaptive computation time
for residual networks \citep{resnet},
which learns the number of residual blocks
required to compute a certain spatial location.
\cite{almahairi16} presented
dynamic capacity networks,
which use the gradient of a coarse output's entropy
to select salient locations
in the input image for refinement.
\cite{ren18} assumed the availability of
\emph{a priori} spatial sparsity in the input image,
and accelerated the convolutional layer
by computing non-sparse regions.

There are dynamic networks that
make binary decisions or multiple choices
for the inference paths taken.
BlockDrop, proposed by \cite{wu18},
trains a policy network to
skip blocks in residual networks.
\cite{liu17d2nn} proposed
conditional branches in deep neural networks (DNNs),
and used Q-learning to train the branching policies.
\cite{odena17} designed a DNN
with layers containing multiple modules,
and decided which module to use
with a recurrent neural network (RNN).
\cite{lin17} learned an RNN to
adaptively prune channels in convolutional layers.
The on/off decisions
commonly used in these networks
cannot be represented by differentiable functions,
hence the gradients are not well-defined.
Consequently, the dynamic networks above
train their policy functions
by reinforcement learning.
There exist, however,
methods that workaround such limitations.
\cite{shazeer17} introduced
sparsely-gated mixture-of-experts
and used a noisy ranking
on the backpropagate-able gating networks
to select the expensive experts
to evaluate.
\cite{bolukbasi17} trained
differentiable policy functions
to implement early exits in a DNN\@.
\cite{hua18} learned
binary policies that decide
whether partial or all input channels
are used for convolution,
but approximate the gradients of
the non-differentiable policy functions
with continuous ones.

%% file: method.tex
\section{Feature Boosting and Suppression}\label{sec:method}

\begin{figure}[ht]
    \includegraphics[width=\textwidth]{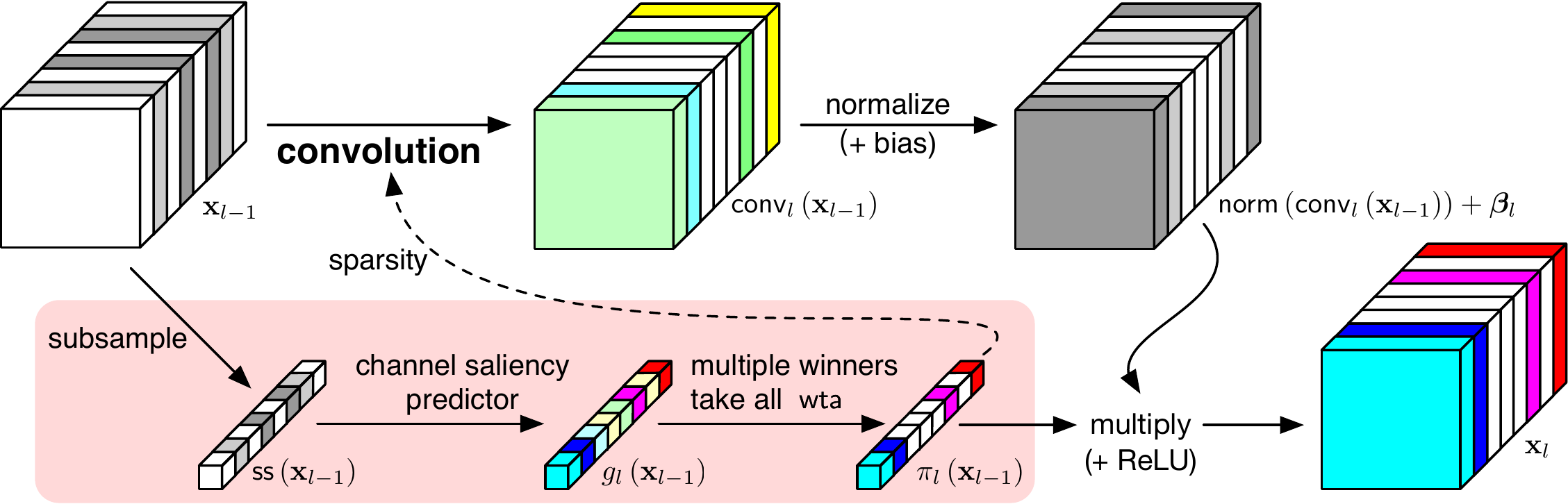}
    \caption{
        A high level view of
        a convolutional layer with FBS.
        By way of illustration,
        we use the $\ordinal{l}$ layer
        with 8-channel input and output features,
        where channels are colored
        to indicate different saliencies,
        and the white blocks (\sparseblock{})
        represent all-zero channels.
    }\label{fig:high_level}
\end{figure}

We start with a high-level illustration
(\Cref{fig:high_level})
of how FBS accelerates a convolutional layer
with \emph{batch normalization} (BN).
The auxiliary components (in red)
predict the importance of each output channel
based on the input features,
and amplify the output features accordingly.
Moreover, certain output channels
are predicted to be entirely suppressed
(or zero-valued as represented by \sparseblock{}),
such output sparsity information
can advise the convolution operation
to skip the computation of these channels,
as indicated by the dashed arrow.
It is notable that the expensive convolution
can be doubly accelerated
by skipping the inactive channels
from both the input features
and the predicted output channel saliencies.
The rest of this section provides
detailed explanation of the components
in \Cref{fig:high_level}.

\subsection{Preliminaries}

For simplicity, we consider
a deep sequential batch-normalized \citep{ioffe15} CNN
with $L$ convolutional layers,
\ie~$
\tx_L =
F(\tx_0) = f_L \left(
    \cdots f_2 \left( f_1 \left( \tx_0 \right) \right) \cdots \right)
$, where the $\ordinal{l}$ layer
$
    f_l: \realset^{C_{l-1} \times H_{l-1} \times W_{l-1}} \to
         \realset^{C_l \times H_l \times W_l}
$
computes the features
$\tx_l \in \realset^{C_l \times H_l \times W_l}$,
which comprise of $C_l$ channels of
features with height $H_l$ and width $W_l$.
The $\ordinal{l}$ layer is thus defined as:
\begin{equation}
    f_l \left( \tx_{l-1} \right) = \left(
        \bnscale_l \cdot \norm \left(
            \conv_l \left( \tx_{l-1}, \convweight_l \right)
        \right) + \bnshift_l
    \right)_+.
    \label{eq:layer}
\end{equation}
Here,
additions ($+$) and multiplications ($\cdot$) are element-wise,
$(\tz)_+ = \max \left( \tz, 0 \right)$ denotes the ReLU activation,
$\bnscale_l, \bnshift_l \in \realset^{C_l}$ are trainable parameters,
$\norm\left(\tz\right)$
normalizes each channel of features $\tz$
across the population of $\tz$,
with $\tmu_\tz, \tsigma_\tz^2 \in \realset^{C_l}$
respectively containing
the population mean and variance of each channel,
and a small $\epsilon$ prevents division by zero:
\begin{equation}
    \norm \left( \tz \right) =
        \frac{\tz - \tmu_\tz}{\sqrt{\tsigma_\tz^2 + \epsilon}}.
\end{equation}
Additionally,
$\conv_l \left( \tx_{l-1}, \convweight_l \right)$
computes the convolution of input features $x_{l-1}$
using the weight tensor
$\convweight_l \in \realset^{C^l \times C^{l-1} \times k^2}$,
where $k$ is the kernel size.
Specifically, FBS concerns
the optimization of
$\conv_l \left( \tx_{l-1}, \convweight_l \right)$ functions,
as a CNN spends
the majority of its inference time in them,
using $k^2 C_{l-1} C_l H_l W_l$
\emph{multiply-accumulate operations} (MACs)
for the $\ordinal{l}$ layer.

\subsection{Designing a Dynamic Layer}%
\label{sec:method:dynamic}

Consider the following generalization
of a layer with dynamic execution:
\begin{equation}
    \hat{f} \left( \tx, \cdots \right) =
        f \left( \tx, \convweight, \cdots \right)
        \cdot
        \pi \left( \tx, \gateweight, \cdots \right),
\end{equation}
where $f$ and $\pi$
respectively use weight parameters
$\convweight$ and $\gateweight$
and may have additional inputs,
and compute tensors of the same output shape,
denoted by $\mathbf{F}$ and $\mathbf{G}$.
Intuitively, the expensive
$\mathbf{F}^{[\ti]}$
can always be skipped for any index $\ti$
whenever the cost-effective
$\mathbf{G}^{[\ti]}$
evaluates to $\mathbf0$.
Here, the superscript $[\ti]$ is used to
index the $\ordinal\ti$ slice of the tensor.
For example, if we have features
$\mathbf{F} \in \realset^{C \times H \times W}$
containing $C$ channels of $H$-by-$W$ features,
$\mathbf{F}^{[c]} \in \realset^{H \times W}$
retrieves the $\ordinal{c}$ feature image.
We can further sparsify and accelerate the layer
by adding, for instance,
a Lasso on $\pi$ to the total loss,
where $\expectation_\tx \left[ \tz \right]$
is the expectation of $\tz$ over $\tx$:
\begin{equation}
    \mathcal{R} \left( \tx \right) =
        \expectation_{\tx} \left[
            {\left\|
                \pi \left( \tx, \gateweight, \cdots \right)
            \right\|}_1
        \right],
    \label{eq:lasso}
\end{equation}
Despite the simplicity of this formulation,
it is however very tricky
to design $\hat{f}$ properly.
Under the right conditions,
we can arbitrarily minimize the Lasso
while maintaining the same output from the layer
by scaling parameters.
For example,
in low-cost collaborative layers \citep{dong17},
$f$ and $\pi$ are simply convolutions
(with or without ReLU activation)
that respectively have weights $\convweight$ and $\gateweight$.
Since $f$ and $\pi$ are homogeneous functions,
one can always halve $\gateweight$ and double $\convweight$
to decrease \eqref{eq:lasso} while
the network output remains the same.
In other words, the optimal network must have
$\left\|\gateweight\right\|_\infty \to 0$,
which is infeasible in finite-precision arithmetic.
For the above reasons,
\cite{dong17} observed that
the additional loss in~\eqref{eq:lasso}
always degrades the CNN's task performance.
\cite{ye18} pointed out that
gradient-based training algorithms
are highly inefficient in exploring
such reparameterization patterns,
and channel pruning methods may
experience similar difficulties.
\cite{shazeer17} avoided this limitation
by finishing $\pi$ with a softmax normalization,
but \eqref{eq:lasso} can no longer be used
as the softmax renders the $\ell^1$-norm,
which now evaluates to 1, useless.
In addition, similar to sigmoid,
softmax (without the cross entropy)
is easily saturated,
and thus may equally
suffer from vanishing gradients.
Many instead design $\pi$ to produce on/off decisions
and train them with reinforcement learning
as discussed in \Cref{sec:related}.

\subsection{Feature Boosting and Suppression with Channel Saliencies}%
\label{sec:method:boosting}

Instead of imposing sparsity on
features or convolutional weight parameters
(\eg~\cite{wen16,alvarez16,li17,he18}),
recent channel pruning methods \citep{liu17, ye18}
induce sparsity on the BN scaling factors $\bnscale_l$.
Inspired by them,
FBS similarly generates
a channel-wise importance measure.
Yet contrary to them,
instead of using the constant BN scaling factors $\bnscale_l$,
we predict channel importance
and dynamically amplify or suppress channels
with a parametric function $\pi(\tx_{l-1})$ dependent on
the output from the previous layer $\tx_{l-1}$.
Here, we propose to
replace the layer definition $f_l \left( \tx_{l-1} \right)$
for each of $l \in [1, L]$ with
$\hat{f}_l \left( \tx_{l-1} \right)$
which employs dynamic channel pruning:
\begin{equation}
    \hat{f}_l \left( \tx_{l-1} \right) = \left(
        \pi_l \left( \tx_{l-1} \right) \cdot
        \left(
            \norm \left(
                \conv_l \left( \tx_{l-1}, \convweight_l \right)
            \right)
            + \bnshift_l
        \right)
    \right)_+,
    \label{eq:fbs}
\end{equation}
where a low-overhead policy $\pi_l \left( \tx_{l-1} \right)$
evaluates the pruning decisions for the computationally demanding
$\conv \left( \tx_{l-1}, \convweight_l \right)$:
\begin{equation}
    \pi_l \left( \tx_{l-1} \right) =
        \wta_{\lceil d C_l \rceil} \left(
            g_l \left( \tx_{l-1} \right)
        \right).
    \label{eq:policy}
\end{equation}
Here, $\wta_k (\tz)$ is  % TODO a figure would be nice.
a $k$-winners-take-all function,
\ie~it returns a tensor identical to $\tz$,
except that we zero out entries
in $\tz$ that are smaller than
the $k$ largest entries in absolute magnitude.
In other words,
$\wta_{\lceil d C_l \rceil} (g_l(\tx_{l-1}))$
provides a pruning strategy that
computes only $\lceil d C_l \rceil$
most salient channels predicted by $g_l(\tx_{l-1})$,
and suppresses the remaining channels with zeros.
In \Cref{sec:method:learn},
we provide a detailed explanation of
how we design a cheap $g_l(\tx_{l-1})$
that learns to predict channel saliencies.

It is notable that our strategy prunes
$C_l - \lceil d C_l \rceil$
least salient output channels from $\ordinal{l}$ layer,
where the density $d \in \left]0, 1\right]$
can be varied to sweep the trade-off relationship
between performance and accuracy.
Moreover, pruned channels contain all-zero values.
This allows the subsequent $\smash{\ordinal{(l+1)}}$ layer
to trivially make use of input-side sparsity,
since all-zero features can be safely skipped
even for zero-padded layers.
Because all convolutions can exploit
both input- and output-side sparsity,
the speed-up gained from pruning is
quadratic with respect to the pruning ratio.
For instance, dynamically pruning
half of the channels in all layers
gives rise to a dynamic CNN
that uses approximately $\frac14$
of the original MACs.

Theoretically, FBS does not
introduce the reparameterization
discussed in \Cref{sec:method:dynamic}.
By batch normalizing the convolution output,
the convolution kernel $\convweight_l$
is invariant to scaling.
Computationally, it is more efficient to train.
Many alternative methods
use non-differentiable $\pi$ functions
that produce on/off decisions.
In general,
DNNs with these policy functions are
incompatible with SGD,
and resort to reinforcement learning for training.
In contrast, \eqref{eq:policy}
allows end-to-end training,
as $\wta$ is a piecewise differentiable
and continuous function like ReLU\@.
\cite{srivastava15} suggested that in general,
a network is easier and faster to train for complex tasks
and less prone to catastrophic forgetting,
if it uses functions such as $\wta$
that promote local competition
between many subnetworks.

\subsection{Learning to Predict Channel Saliencies}\label{sec:method:learn}

This section explains the design of
the channel saliency predictor $g_l (\tx_{l-1})$.
To avoid significant computational cost in $g_l$,
we subsample $\tx_{l-1}$ by
reducing the spatial dimensions
of each channel to a scalar
using the following function
$\subs: \realset^{C \times H \times W} \to \realset^C$:
\begin{equation}
    \subs \left( \tx_{l-1} \right) =
        \frac{1}{HW} \left[
            \subsubs \left( \tx_{l-1}^{[1]} \right) ~
            \subsubs \left( \tx_{l-1}^{[2]} \right) ~
            \cdots ~
            \subsubs \left( \tx_{l-1}^{[C]} \right)
        \right],
\end{equation}
where $\subsubs \left( \tx_{l-1}^{[c]} \right)$
reduces the $\ordinal{c}$ channel of $\tz$ to a scalar
using, for instance,
the $\ell^1$-norm $\smash{\| \tx_{l-1}^{[c]} \|_1}$,
$\ell^2$-norm, $\ell^\infty$-norm,
or the variance of $\smash{\tx_{l-1}^{[c]}}$.
The results in \Cref{sec:results}
use the $\ell^1$-norm by default,
which is equivalent to global average pooling
for the ReLU activated $\tx_{l-1}$.
We then design $g_l (\tx_{l-1})$ to
predict channel saliencies
with a fully connected layer
following the subsampled activations
$\subs\left( \tx_{l-1} \right)$,
where $\gateweight_l \in \realset^{C^l \times C^{l-1}}$
is the weight tensor of the layer:
\begin{equation}
    g_l \left( \tx_{l-1} \right) = \left(
        \subs\left( \tx_{l-1} \right) \gateweight_l + \gatebias_l
    \right)_+.
    \label{eq:our_policy}
\end{equation}
We generally initialize $\gatebias_l$ with $1$
and apply \cite{he15}'s initialization to $\gateweight_l$.
Similar to how \cite{liu17} and \cite{ye18}
induced sparsity in the BN scaling factors,
we regularize all layers
with the Lasso on $g_l (\tx_{l-1})$:
$
    \lambda \sum_{l=1}^L
        \expectation_{\tx} \left[
            {\left\|g_l(\tx_{l-1})\right\|}_1
        \right]
$
in the total loss,
where $\lambda = 10^{-8}$ in our experiments.

%% file: results.tex
\section{Experiments}\label{sec:results}

We ran extensive experiments
on \cifarx{} \citep{cifar}
and the ImageNet \ilsvrc{} \citep{imagenet},
two popular image classification datasets.
We use \mcnet{} \citep{mayo},
a custom 8-layer CNN for \cifarx{}
(see \Cref{app:cifarx} for its structure),
using only $1.3\unitM$ parameters
with $91.37\%$ and $99.67\%$
top-1 and top-5 accuracies respectively.
\mcnet{} is much smaller
than a \vggxvi{} on \cifarx{} \citep{liu17},
which uses $20\unitM$ parameters
and only $2.29\%$ more accurate.
Because of its compactness,
our CNN is more challenging to accelerate.
By faithfully reimplementing
\emph{Network Slimming} (NS) \citep{liu17},
we closely compare FBS with NS
under various speedup constraints.
For \ilsvrc, we augment two popular CNN variants,
\resnetxviii{} \citep{resnet} and \vggxvi{} \citep{vgg},
and provide detailed accuracy/MACs
trade-off comparison against
recent structured pruning
and dynamic execution methods.

Our method begins by first replacing
all convolutional layer computations with \eqref{eq:fbs},
and initializing the new convolutional kernels
with previous parameters.
Initially, we do not suppress any channel computations
by using density $d = 1$ in \eqref{eq:policy}
and fine-tune the resulting network.
For fair comparison against NS,
we then follow \cite{liu17}
by iteratively decrementing
the overall density $d$ of the network
by $10\%$ in each step,
and thus gradually using fewer channels
to sweep the accuracy/performance trade-off.
The difference is that
NS prunes channels by ranking globally,
while FBS prunes around $1 - d$ of each layer.

\subsection{\cifarx}\label{sec:results:cifar10}

\begin{figure}[ht]
    \centering%
    \begin{subfigure}[t]{0.45\textwidth}
        \includegraphics[width=0.8\textwidth]{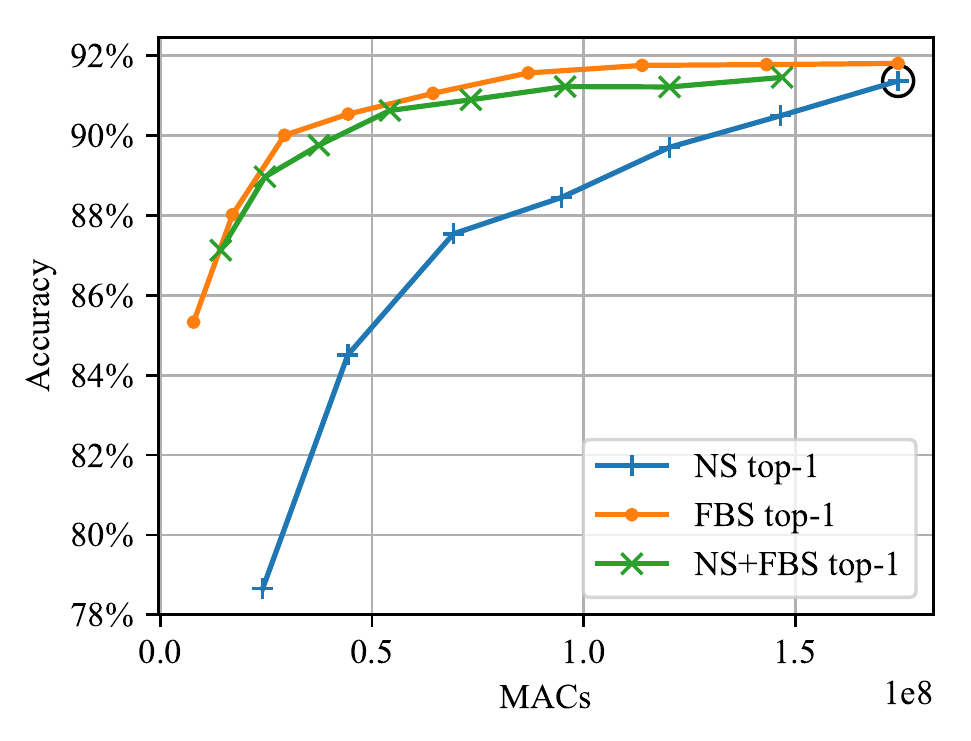}%
        \vspace{-5pt}%
        \caption{%
            \mcnet{} accuracy/MACs trade-off
        }\label{fig:cifar10:compare}
    \end{subfigure}\quad
    \begin{subfigure}[t]{0.45\textwidth}
        \includegraphics[width=0.8\textwidth]{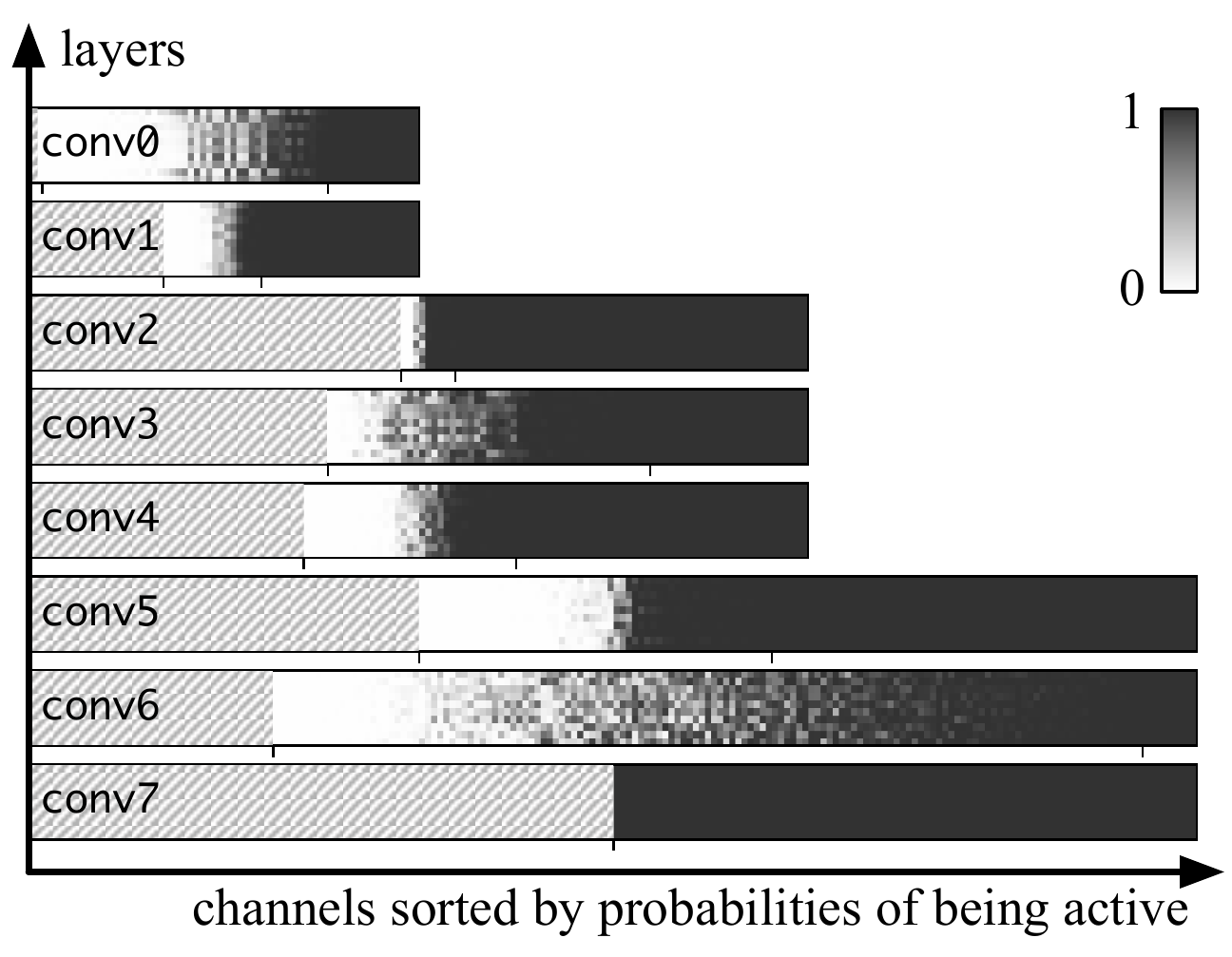}%
        \vspace{-5pt}%
        \caption{%
            Channel skipping probabilites
        }\label{fig:cifar10:heatmap}
    \end{subfigure}%
    \caption{%
        Experimental results on \mcnet{}.
        We compare
        in (\subref{fig:cifar10:compare})
        the accuracy/MACs trade-off
        between FBS, NS and FBS+NS\@.
        The baseline is emphasized
        by the circle $\bigcirc$.
        The heat map in (\subref{fig:cifar10:heatmap})
        reveals the individual probability
        of skipping a channel
        for each channel ($x$-axis),
        when an image of a category ($y$-axis)
        is shown to the network with $d = 1$.
    }\label{fig:cifar10}\vspace{-9pt}%
\end{figure}

% \tikzset{
%     cross/.style={
%         cross out, draw,
%         minimum size=2*(#1-\pgflinewidth),
%         inner sep=0pt, outer sep=0pt,
%         line width=0.2mm,
%     },
%     cross/.default={3pt}
% }
% \definecolor{ns}{RGB}{38,120,178}
% \newcommand{\nslegend}{\tikz{
%     \draw [ns, line width=0.3mm]
%         (-0.3, 0) -- (0.3, 0);
%     \draw [ns, line width=0.2mm]
%         (0, -0.1) -- (0, 0.1);
% }}
% \definecolor{fbs}{RGB}{253,127,40}
% \newcommand{\fbslegend}{\tikz{
%     \draw (0, 0) node[scale=0.3, shape=circle, fill=fbs] {};
%     \draw [fbs, line width=0.3mm] (-0.3, 0) -- (0.3, 0);
% }}
% \definecolor{nsfbs}{RGB}{51,159,52}
% \newcommand{\nsfbslegend}{\tikz{
%     \draw (0, 0) node[nsfbs, cross] {};
%     \draw [nsfbs, line width=0.3mm]
%         (-0.3, 0) -- (0.3, 0);
% }}

By respectively applying NS and FBS
to our \cifarx{} classifier
and incrementally increasing sparsity,
we produce the trade-off relationships
between number of operations (measured in MACs)
and the classification accuracy
as shown in \Cref{fig:cifar10:compare}.
FBS clearly surpasses NS
in its ability to retain the task accuracy
under an increasingly stringent computational budget.
Besides comparing FBS against NS,
we are interested in combining both methods,
which demonstrates the effectiveness of FBS
if the model is already less redundant,
\ie~it cannot be pruned further using NS
without degrading the accuracy by more than $1\%$.
The composite method (NS+FBS)
is shown to successfully regain
most of the lost accuracy due to NS,
producing a trade-off curve closely matching FBS\@.
It is notable that
under the same $90.50\%$ accuracy constraints,
FBS, NS+FBS, and NS respectively achieve
$3.93\times$, $3.22\times$, and $1.19\times$
speed-up ratios.
Conversely for a $2\times$ speed-up target,
they respectively produce models
with accuracies not lower than
$91.55\%$, $90.90\%$ and $87.54\%$.

\Cref{fig:cifar10:heatmap} demonstrates that
our FBS can effectively learn
to amplify and suppress channels
when dealing with different input images.
The 8 heat maps respectively
represent the channel skipping probabilities
of the 8 convolutional layers.
The brightness of the pixel
at location $(x, y)$ denotes
the probability of skipping
the $\ordinal{x}$ channel
when looking at an image
of the $\ordinal{y}$ category.
The heat maps verify our belief that
the auxiliary network learned to predict
which channels specialize to which features,
as channels may have drastically distinct
probabilites of being used
for images of different categories.
The model here is a \mcnet{}
using FBS with $d = 0.5$,
which has a top-1 accuracy of
$90.59\%$ (top-5 $99.65\%$).
Moreover, channels in the heat maps are sorted
so the channels that are
on average least frequently evaluated
are placed on the left,
and channels shaded in stripes are never evaluated.
The network in \Cref{fig:cifar10:heatmap}
is not only approximately $4\times$ faster
than the original,
by removing the unused channels,
we also reduce the number of weights by $2.37\times$.
This reveals that FBS naturally subsumes
channel pruning strategies such as NS,
as we can simply prune away channels
that are skipped regardless of the input.
It is notable that even though we specified
a universal density $d$,
FBS learned to
adjust its dynamicity across all layers,
and prune different ratios of channels
from the convolutional layers.

\subsection{ImageNet \ilsvrc{} Classification}\label{sec:results:resnet}

By applying FBS and NS respectively to \resnetxviii{},
we saw that the \ilsvrc{} validation accuracy
of FBS consistently outperforms NS
under different speed-up constraints
(see \Cref{app:ilsvrc} for
 the implementation details and trade-off curves).
For instance, at $d = 0.7$,
it utilizes only $1.12\unitG$ MACs
($1.62\times$ fewer)
to achieve a top-1 error rate of $31.54\%$,
while NS requires $1.51\unitG$ MACs
($1.21\times$ fewer)
for a similar error rate of $31.70\%$.
When compared across
recent dynamic execution methods
examined in \Cref{tab:compare:resnet},
FBS demonstrates simultaneously
the highest possible speed-up
and the lowest error rates.
It is notable that
the baseline accuracies for FBS refer to
a network that has been augmented
with the auxiliary layers featuring FBS
but suppress no channels, \ie~$d = 1$.
We found that this method brings immediate
accuracy improvements,
an increase of $1.73\%$ in top-1
and $0.46\%$ in top-5 accuracies,
to the baseline network,
which is in line with our observation on \mcnet.

In \Cref{tab:compare:vgg},
we compare different structured pruning
and dynamic execution methods to FBS for \vggxvi{}
(see \Cref{app:ilsvrc} for the setup).
At a speed-up of $3.01\times$,
FBS shows a minimal increase
of $0.44\%$ and $0.04\%$
in top-1 and top-5 errors respectively.
At $5.23\times$ speed-up,
it only degrades
the top-1 error by $1.08\%$
and the top-5 by $0.59\%$.

Not only does FBS use much fewer MACs,
it also demonstrates significant reductions
in bandwidth and memory requirements.
In \Cref{tab:compare:mem},
we observe a large reduction
in the number of memory accesses
in single image inference
as we simply do not access
suppressed weights and activations.
Because these memory operations
are often costly DRAM accesses,
minimizing them leads to power-savings.
\Cref{tab:compare:mem} further reveals that
in diverse application scenarios
such as low-end and cloud environments,
the peak memory usages by the optimized models
are much smaller than the originals,
which in general improves cache utilization.

\begin{table*}[ht]
    \centering%
    \begin{adjustbox}{scale=0.8}
    \begin{tabular}{lcrrrrr}
        \toprule
        \multirow{2}{*}{\textbf{Method}}
            & \hspace{-15pt}
              \multirow{2}{*}{\textbf{Dynamic}}
              \hspace{-5pt}
            & \multicolumn{2}{c}{\textbf{Baseline}}
            & \multicolumn{2}{c}{\textbf{Accelerated}}
            & \textbf{MAC} \\
        ~
            & ~
            & Top-1
            & Top-5
            & Top-1
            & Top-5
            & \textbf{saving} \\
        \midrule \midrule
        \emph{Soft Filter Pruning} \citep{he18}
            &
            & 29.72
            & 10.37
            & 32.90
            & 12.22
            & 1.72$\times$ \\
        \emph{Network Slimming} (\cite{liu17}, our implementation)
        \hspace{-20pt}
            &
            & 31.02
            & 11.32
            & 32.79
            & 12.61
            & 1.39$\times$ \\
        \emph{Discrimination-aware Channel Pruning} \citep{zhuang18}
        \hspace{-20pt}
            &
            & 30.36
            & 11.02
            & 32.65
            & 12.40
            & 1.89$\times$ \\
        \emph{Low-cost Collaborative Layers} \citep{dong17}
            & \checkmark
            & 30.02
            & 10.76
            & 33.67
            & 13.06
            & 1.53$\times$ \\
        \emph{Channel Gating Neural Networks} \citep{hua18}
            & \checkmark
            & 30.98
            & 11.16
            & 32.60
            & 12.19
            & 1.61$\times$ \\
        \rowcolor{Gray}
        \emph{Feature Boosting and Suppression} (FBS)
            & \checkmark
            & \textbf{29.29}
            & \textbf{10.32}
            & \textbf{31.83}
            & \textbf{11.78}
            & \textbf{1.98$\times$} \\
        \bottomrule
    \end{tabular}
    \end{adjustbox}%
    \vspace{-8pt}%
    \caption{%
        Comparisons of error rates
        of the baseline and accelerated \resnetxviii{} models.
    }\label{tab:compare:resnet}
\end{table*}%
\vspace{-10pt}%
\begin{table*}[!ht]
    \centering
    \begin{adjustbox}{scale=0.8}
    \begin{tabular}{lcrrr}
        \toprule
        \multirow{2}{*}{\textbf{Method}}
            & \hspace{-15pt}
              \multirow{2}{*}{\textbf{Dynamic}}
              \hspace{-5pt}
            & \multicolumn{3}{c}{\textbf{$\Delta$ top-5 errors (\%)}} \\
        {}
            &
            & 3$\times$
            & \hspace{13pt}4$\times$
            & 5$\times$ \\
        \midrule \midrule
        \emph{Filter Pruning}
        (\cite{li17}, reproduced by \cite{he17})
            &
            & ---
            & 8.6
            & 14.6 \\
        \emph{Perforated CNNs} \citep{figurnov16}
            &
            & 3.7
            & 5.5
            & --- \\
        \emph{Network Slimming} (\cite{liu17}, our implementation)
            &
            & 1.37
            & 3.26
            & 5.18 \\
        \emph{Runtime Neural Pruning} \citep{lin17}
            & \checkmark
            & 2.32
            & 3.23
            & 3.58 \\
        \emph{Channel Pruning} \citep{he17}
            &
            & 0.0
            & 1.0
            & 1.7 \\
        \emph{AutoML for Model Compression} \citep{he18amc}
            &
            & ---
            & ---
            & 1.4 \\
        \emph{ThiNet-Conv} \citep{luo17}
            &
            & 0.37
            & ---
            & --- \\
        \rowcolor{Gray}
        \emph{Feature Boosting and Suppression} (FBS)
            & \checkmark
            & 0.04
            & \textbf{0.52}
            & \textbf{0.59} \\
        \bottomrule
    \end{tabular}
    \end{adjustbox}%
    \vspace{-8pt}%
    \caption{%
        Comparisons of top-5 error rate increases
        for \vggxvi{} on \ilsvrc{} validation set
        under 3$\times$, 4$\times$ and 5$\times$
        speed-up constraints.
        The baseline has a $10.1\%$ top-5 error rate.
        Results from \cite{he17} only show numbers
        with one digit after the decimal point.
    }\label{tab:compare:vgg}
\end{table*}%
\vspace{-10pt}%
\begin{table*}[!ht]
    \centering
    \begin{adjustbox}{scale=0.8}
    \begin{tabular}{lllll}
        \toprule
        \multirow{2}{*}{\textbf{Model}}
            & \multicolumn{2}{c}{\textbf{%
                Total Memory Accesses}}
            & \multicolumn{2}{c}{\textbf{Peak Memory Usage}} \\
        {}
            & {Weights}
            & {Activations}
            & {Edge (1 image)}
            & {Cloud (128 images)} \\
        \midrule \midrule
        \emph{\vggxvi{}}
            & $56.2$ MB
            & $86.5$ MB
            & $24.6$ MB
            & $3.09$ GB \\
        \emph{\vggxvi{} $3\times$}
            & $23.9$ MB ($2.35 \times$)
            & $40.8$ MB ($2.12 \times$)
            & $9.97$ MB ($2.47 \times$)
            & $1.24$ GB ($2.47 \times$) \\
        \midrule
        \emph{\resnetxviii{}}
            & $44.6$ MB
            & $17.8$ MB
            & $9.19$ MB
            & $0.47$ GB \\
        \emph{\resnetxviii{} $2\times$}
            & $20.5$ MB ($2.18 \times$)
            & $12.3$ MB ($1.45 \times$)
            & $4.68$ MB ($1.96 \times$)
            & $0.31$ GB ($1.49 \times$) \\
        \bottomrule
    \end{tabular}
    \end{adjustbox}%
    \vspace{-8pt}%
    \caption{%
        Comparisons of the memory accesses
        and peak memory usage
        of the \ilsvrc{} classifiers with FBS
        respectively under $3\times$ and $2\times$
        inference speed-ups.
        The \textbf{Weights} and \textbf{Activations}
        columns respectively show
        the total amount of weight and activation
        accesses required by all convolutions
        for a single image inference.
        The \textbf{Peak Memory Usage} columns
        show the peak memory usages
        with different batch sizes.
    }\label{tab:compare:mem}
\end{table*}

%% file: conclusion.tex
\section{Conclusion}\label{sec:conclusion}

In summary,
we proposed feature boosting and suppression
that helps CNNs to achieve
significant reductions in the compute required
while maintaining high accuracies.
FBS fully preserves the capabilities of CNNs
and predictively boosts important channels
to help the accelerated models retain high accuracies.
We demonstrated that
FBS achieves around $2\times$ and $5\times$ savings in computation
respectively on \resnetxviii{} and \vggxvi{}
within $0.6\%$ loss of top-5 accuracy.
Under the same performance constraints,
the accuracy gained by FBS surpasses
all recent structured pruning and dynamic execution methods
examined in this paper.
In addition,
it can serve as an off-the-shelf technique
for accelerating many popular CNN networks
and the fine-tuning process
is unified in the traditional SGD
which requires no algorithmic changes in training.
Finally, the implementation of FBS
and the optimized networks are
fully open source and released to the public%
\footnote{\url{https://github.com/deep-fry/mayo}}.

%% file: acknowledgements.tex
\section*{Acknowledgements}

This work is supported in part by
the National Key R\&D Program of China (No\@. 2018YFB1004804),
the National Natural Science Foundation of China (No\@. 61806192).
We thank EPSRC for providing Yiren Zhao
his doctoral scholarship.

%% file: appendix.tex
\section{Details of \mcnet{} on \cifarx{}}\label{app:cifarx}

For the \cifarx{} classification task,
we use \mcnet, a custom designed CNN,
with less than $1.30\unitM$ parameters
and takes $174\unitM$ MACs
to perform inference
for a $32$-by-$32$ RGB image.
The architecture is illustrated
in \Cref{tab:mcnet},
where all convolutional layers
use $3 \times 3$ kernels,
the \textbf{Shape} column shows
the shapes of each layer's features,
and \texttt{pool7} is a global average pooling layer.

We trained \mcnet{} (see \Cref{app:cifarx})
with a $0.01$ learning rate
and a $256$ batch size.
We reduced the learning rate
by a factor of $10\times$ for every $100$ epochs.
To compare FBS against NS fairly,
every model with a new target MACs budget
were consecutively initialized with the previous model,
and trained for a maximum of $300$ epochs,
which is enough for all models
to converge to the best obtainable accuracies.
For NS, we follow \cite{liu17}
and start training
with an $\ell^1$-norm sparsity regularization
weighted by $10^{-5}$
on the BN scaling factors.
We then prune at $150$ epochs
and fine-tune the resulting network
without the sparsity regularization.

We additionally employed image augmentation
procedures from \cite{alexnet}
to preprocess each training example.
Each \cifarx{} example was randomly horizontal flipped
and slightly perturbed in the brightness, saturation and hue.

\Cref{tab:mcnet} additionally provides
further comparisons
of layer-wise compute costs between FBS, NS,
and the composition of the two methods (NS+FBS).
It is notable that the FBS column
has two different output channel counts,
where the former is the number of computed channels
for each inference,
and the latter is the number of channels
remaining in the layer
after removing the unused channels.
\begin{table}[ht]
    \centering
        \begin{tabular}{llrrrr}
            \toprule
            \textbf{Layer}
                & \textbf{Shape}
                & \multicolumn{4}{c}{\textbf{Number of MACs (Output Channels)}} \\
            ~
                & ~
                & Original
                & NS
                & FBS
                & NS+FBS \\
            \midrule\midrule
            \texttt{conv0}
                & $30 \times 30$
                & 1.5\unitM~(64)
                & 1.3\unitM~(52)
                & 893\unitk~(32/62)
                & 860\unitk~(32) \\
            \texttt{conv1}
                & $30 \times 30$
                & 33.2\unitM~(64)
                & 27.0\unitM~(64)
                & 8.4\unitM~(32/42)
                & 10.2\unitM~(39) \\
            \texttt{conv2}
                & $15 \times 15$
                & 16.6\unitM~(128)
                & 15.9\unitM~(123)
                & 4.2\unitM~(64/67)
                & 5.9\unitM~(74) \\
            \texttt{conv3}
                & $15 \times 15$
                & 33.2\unitM~(128)
                & 31.9\unitM~(128)
                & 8.3\unitM~(64/79)
                & 11.6\unitM~(77) \\
            \texttt{conv4}
                & $15 \times 15$
                & 33.2\unitM~(128)
                & 33.1\unitM~(128)
                & 8.3\unitM~(64/83)
                & 12.1\unitM~(77) \\
            \texttt{conv5}
                & $8 \times 8$
                & 14.1\unitM~(192)
                & 13.4\unitM~(182)
                & 3.6\unitM~(96/128)
                & 4.9\unitM~(110) \\
            \texttt{conv6}
                & $8 \times 8$
                & 21.2\unitM~(192)
                & 11.6\unitM~(111)
                & 5.4\unitM~(96/152)
                & 4.3\unitM~(67) \\
            \texttt{conv7}
                & $8 \times 8$
                & 21.2\unitM~(192)
                & 12.3\unitM~(192)
                & 5.4\unitM~(96/96)
                & 4.5\unitM~(116) \\
            \texttt{pool7}
                & $1 \times 1$ & & & & \\
            \texttt{fc}
                & $1 \times 1$
                & 1.9\unitk~(10)
                & 1.9\unitk~(10)
                & 960~(10)
                & 1.1\unitk~(10) \\
            \midrule
            Total
                &
                & 174.3\unitM
                & 146.5\unitM
                & 44.3\unitM
                & 54.2\unitM \\
            Saving
                &
                & -
                & $1.19\times$
                & $3.93\times$
                & $3.21\times$ \\
            \bottomrule
        \end{tabular}
        \caption{%
            The network structure of \mcnet{}
            for \cifarx{} classification.
            In addition, we provide a detailed
            per-layer MACs comparison between
            FBS, NS, and the composition of them (NS+FBS).
            We minimize the models generated
            by the three methods
            while maintaining a classification accuracy
            of at least $90.5\%$.
        }\label{tab:mcnet}
\end{table}

\begin{figure}[!ht]
    \centering%
    \includegraphics[width=0.5\textwidth]{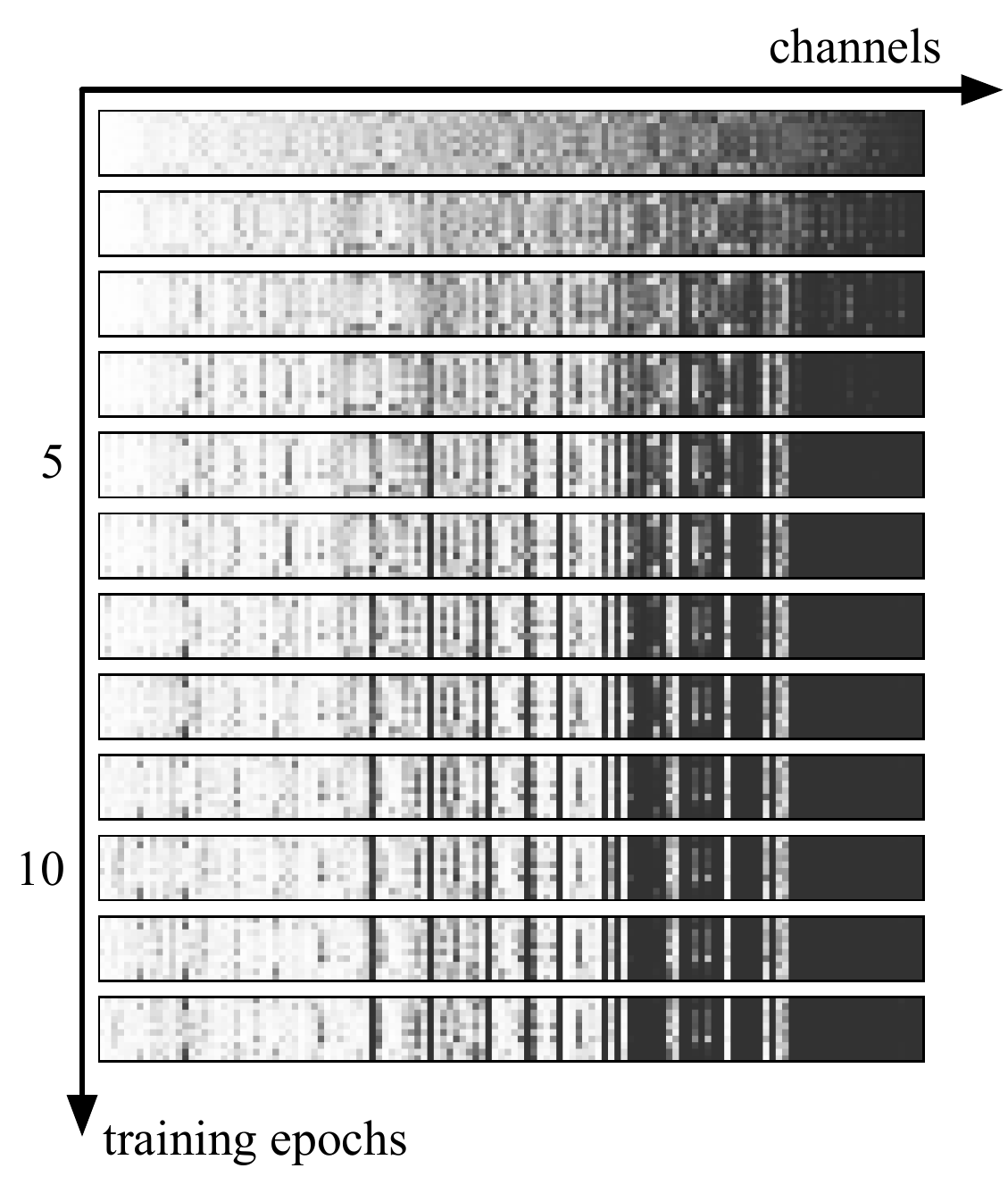}
    \caption{%
        The training history
        of a convolutional layer \texttt{conv4}
        in \mcnet.
        The history is visualized
        by the 12 skipping probabilites heat maps,
        where the heights denote the 10 categories in \cifarx,
        and channels in \texttt{conv4} occupy the width.
    }\label{fig:cifar10:history}
\end{figure}

\Cref{fig:cifar10:history}
shows how the skipping probabilites heat maps
of the convolutional layer \texttt{conv4}
evolve as we fine-tune FBS-augmented \mcnet{}.
The network was trained for 12 epochs,
and we saved the model at every epoch.
The heat maps are generated with the saved models in sequence,
where we apply the same reordering to all heat map channels
with the sorted result from the first epoch.
It can be observed that as we train the network,
the channel skipping probabilites become more pronounced.

\section{Details of the \ilsvrc{} classifiers}\label{app:ilsvrc}

\ilsvrc{} classifiers,
\ie~\resnetxviii{} and \vggxvi{},
were trained with a procedure similar to \Cref{app:cifarx}.
The difference was that
they were trained for a maximum of $35$ epochs,
the learning rate was decayed for every $20$ epochs,
and NS models were all pruned at $15$ epochs.
For image preprocessing, we additionally
cropped and stretched/squeezed images randomly
following \cite{alexnet}.

Since \vggxvi{} is computationally intensive
with over $15\unitG$ MACs,
We first applied NS on \vggxvi{}
to reduce the computational and memory requirements,
and ease the training of the FBS-augmented variant.
We assigned a $1\%$ budget
in top-5 accuracy degradation
and compressed the network using NS\@,
which gave us a smaller \vggxvi{}
with $20\%$ of all channels pruned.
The resulting network is a lot less redundant,
which almost halves the compute requirements,
with only $7.90\unitG$ MACs remaining.
We then apply FBS to the well-compressed network.

Residual networks \citep{resnet}, such as \resnetxviii,
adopt sequential structure of residual blocks:
$
    \tx_b =
        K\left(\tx_{b-1}\right) +
        F\left( \tx_{b-1} \right)
$,
where $\tx_b$ is the output
of the $\ordinal{b}$ block,
$K$ is either an identity function
or a downsampling convolution,
and $F$ consists of a sequence of convolutions.
For residual networks,
we directly apply FBS
to all convolutional layers,
with a difference in the way
we handle the feature summation.
Because the $\ordinal{(b+1)}$ block receives as input
the sum of the two features with sparse channels
$K\left( \tx_{b-1} \right)$ and $F\left( \tx_{b-1} \right)$,
a certain channel of this sum is treated as sparse
when the same channels in both features are simultaneously sparse.

\begin{figure}[!ht]
    \centering%
    % \begin{subfigure}[t]{0.45\textwidth}
        \includegraphics[width=0.7\textwidth]{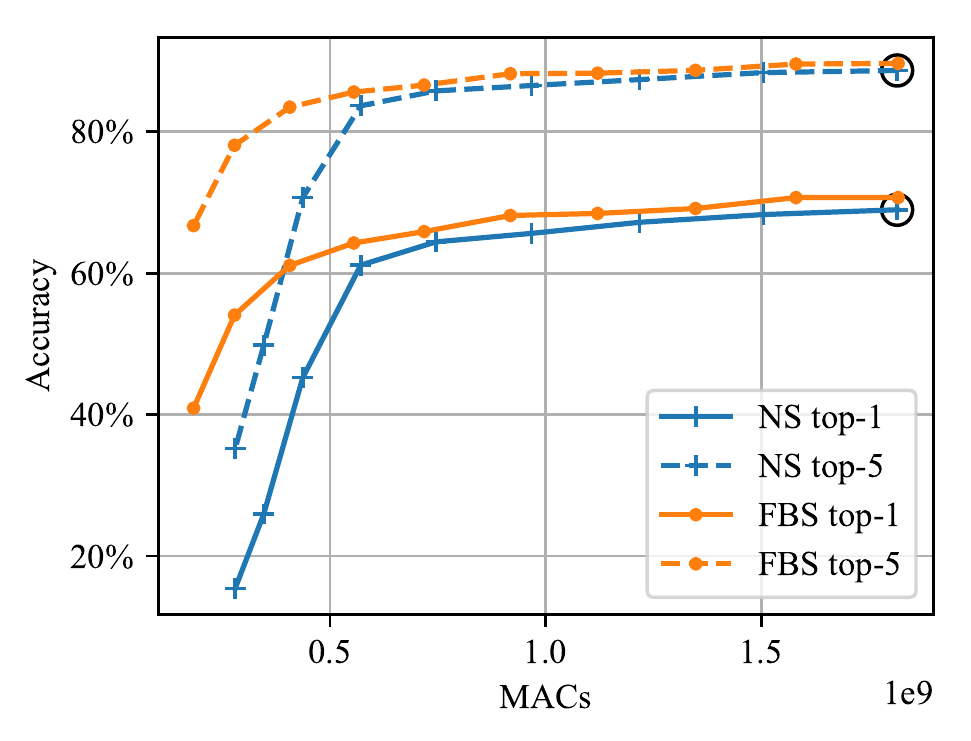}
    %    \caption{%
    %        \resnetxviii{} accuracy/MACs trade-off
    %    }\label{fig:resnet:macs}
    % \end{subfigure}\quad
    % \begin{subfigure}[t]{0.45\textwidth}
    %     \includegraphics[width=\textwidth]{plots/resnet18_speedup.pdf}
    %     \caption{%
    %         Increase in error \emph{vs.} speed-up ratio
    %     }\label{fig:resnet:speedup}
    % \end{subfigure}\quad
    \caption{%
        The accuracy/performance trade-off
        comparison between NS and FBS for \resnetxviii{}
        on the ImageNet \ilsvrc{} validation set.
    }\label{fig:resnet}
\end{figure}

\Cref{fig:resnet} compares
the accuracy/performance trade-off curves
between FBS and NS for \resnetxviii{}.
% \Cref{fig:resnet:speedup} shows the same comparison, but instead use speed-up
